\documentclass[conference]{IEEEtran}
\IEEEoverridecommandlockouts
\usepackage{cite}
\usepackage{amsmath,amssymb,amsfonts}
\usepackage{algorithmic}
\usepackage{graphicx}
\usepackage{textcomp}
\usepackage{xcolor}
\usepackage[colorlinks]{hyperref}

\usepackage{mathtools}
\usepackage{adjustbox}
\def\BibTeX{{\rm B\kern-.05em{\sc i\kern-.025em b}\kern-.08em
    T\kern-.1667em\lower.7ex\hbox{E}\kern-.125emX}}

\DeclareMathOperator*{\OPLUS}{\oplus}

\begin{document}

\title{MR-STGN: Multi-Residual Spatio Temporal Graph Network using Attention Fusion for Patient Action Assessment}

\author{

\IEEEauthorblockN{1\textsuperscript{st} Youssef Mourchid}

\IEEEauthorblockA{
\textit{CESI LINEACT, UR 7527} \\
\textit{Dijon, 21800, France}\\
ymourchid@cesi.fr}
\and
\IEEEauthorblockN{2\textsuperscript{nd} Rim Slama}

\IEEEauthorblockA{
\textit{CESI LINEACT, UR 7527} \\
\textit{Lyon, 69100, France}\\
rsalmi@cesi.fr}

}

\maketitle

\begin{abstract}
Accurate assessment of patient actions plays a crucial role in healthcare as it contributes significantly to disease progression monitoring and treatment effectiveness. However, traditional approaches to assess patient actions often rely on manual observation and scoring, which are subjective and time-consuming. In this paper, we propose an automated approach for patient action assessment using a Multi-Residual Spatio Temporal Graph Network (MR-STGN) that incorporates both angular and positional 3D skeletons. The MR-STGN is specifically designed to capture the spatio-temporal dynamics of patient actions. It achieves this by integrating information from multiple residual layers, with each layer extracting features at distinct levels of abstraction. Furthermore, we integrate an attention fusion mechanism into the network, which facilitates the adaptive weighting of various features. This empowers the model to concentrate on the most pertinent aspects of the patient's movements, offering precise instructions regarding specific body parts or movements that require attention. Ablation studies are conducted to analyze the impact of individual components within the proposed model. 
We evaluate our model on the UI-PRMD dataset demonstrating its performance in accurately predicting real-time patient action scores, surpassing state-of-the-art methods. 
\end{abstract}

\begin{IEEEkeywords}
Automatic assessment, Rehabilitation, Spatio-Temporal, Graph convolution networks, Attention mechanism
\end{IEEEkeywords}

\section{Introduction}
\label{sec:intro}
Physical rehabilitation is a critical component of patient care, particularly for individuals with disabilities or injuries ~\cite{jang2019pulmonary}. However, ensuring patients are performing exercises correctly and adhering to their rehabilitation program can be challenging for healthcare providers ~\cite{kitzman2021physical}. Recent advances in technology have opened up the possibility of using machine learning algorithms to automatically assess patients' performance of physical rehabilitation exercises. However, there are several challenges to overcome in order to achieve reliable and accurate assessment, including sensor placement, data variability, and patient heterogeneity. 

\begin{figure}[htp]
\centering
\includegraphics[width=9cm,height=4cm]{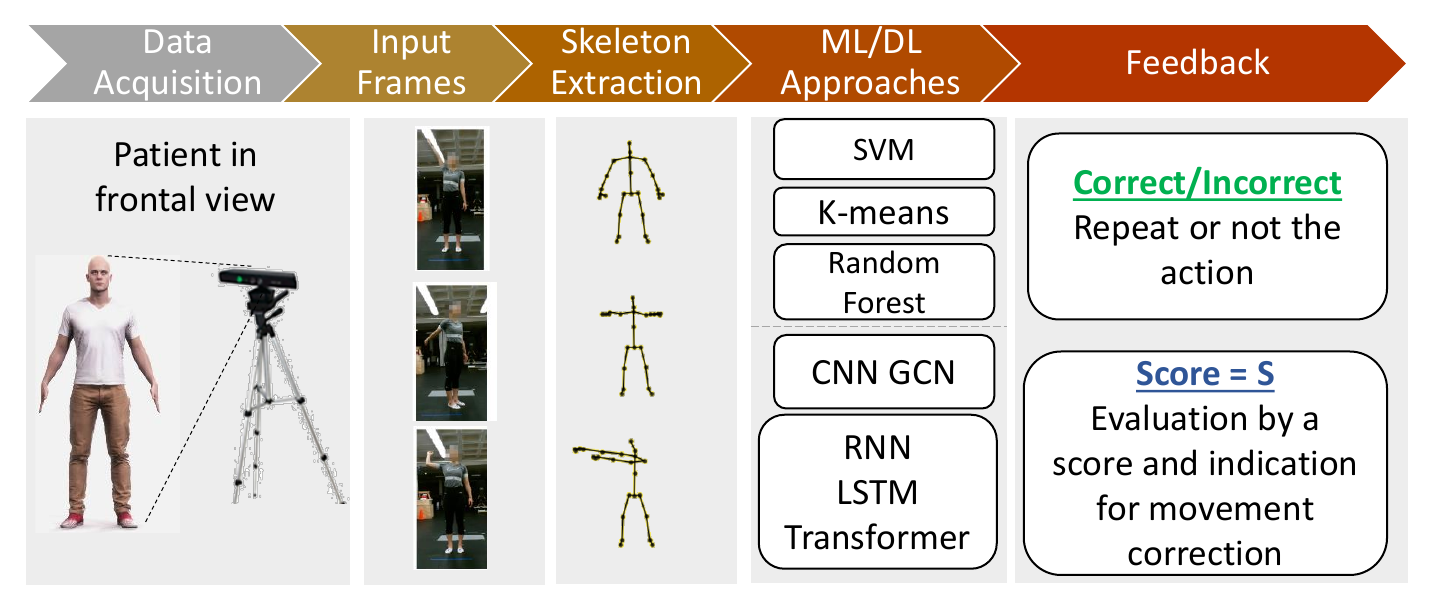}
\caption{Physical rehabilitation exercises process overview. } 
\label{fig:RehabilitationOverview}
\end{figure}

Various technologies, including inertial measurement units (IMUs), computer vision, and machine learning algorithms, have been explored for automatically assessing physical rehabilitation exercises ~\cite{du2021assessing}. IMUs are small sensors attached to a patient's body that measure movement and orientation, while computer vision techniques track joint positions and movements from video footage. Advancements in computer vision, driven by graphs, statistical techniques, and deep learning, have greatly enhanced visual data processing \cite{mourchid2016image,benallal2022new,mourchid2021automatic,lafhel2021movie}. Machine learning algorithms ~\cite{burns2018shoulder}, like artificial neural networks, help recognizing exercise patterns and providing personalized feedback on patient performance using sensor data or video footage. While these approaches show promise in accurately identifying exercises and detecting errors, challenges remain in developing reliable and robust algorithms that can handle data variability and patient heterogeneity. More extensive validation studies in clinical settings are necessary to ensure the safety and effectiveness of these automated systems. Moreover, these approaches consider the exercise evaluation as a binary classification task, distinguishing exercises as either correct or incorrect. Nevertheless, these investigations were constrained in their capacity to furnish comprehensive feedback for individual exercise performances, as depicted in Fig. \ref{fig:RehabilitationOverview}. 

Deep learning approaches ~\cite{panwar2019rehab,zhang2020rehabilitation,liao2020deep} , including CNNs, RNNs, and GCNs, have received attention in the physical rehabilitation field for their ability to automatically assess patients' exercise performance by analyzing sensor data or video footage. The use of Graph convolutional networks (GCNs), which operates on data represented as graphs, offers a unique advantage in capturing spatiotemporal characteristics of exercises and modeling complex interactions between body parts while respecting the natural topological structure of the human body skeleton. However, these approaches mostly require time-consuming processing and need to be improved for better precision. Furthermore, they do not offer a reliable and complete feedback, which can be very helpful for patient exercise improvement.

Inspired by Graph Convolutional Networks (GCNs), we propose a novel Multi-Residual Spatio Temporal Graph Network (MR-STGN) that employs both information from positional and angular 3D skeletons to capture the spatial and temporal dependencies of human body parts. GCNs have shown promising results in modeling graph-structured data, and we leverage their benefits in our proposed model. The use of 3D skeletons allows us to capture the positional and angular information of human body parts, which is crucial in assessing patient actions during rehabilitation exercises. We employ a multi-residual framework to learn deeper representations and mitigate the problem of vanishing gradients in deep neural networks. Additionally, our proposed model utilizes a graph attention mechanism that selectively attends to relevant nodes and edges in the graph, improving the model's discriminative power. Furthermore, our model provides clear guidance on which body parts or movements to focus on, leading to improved assessment quality. The real-time performance of the MR-STGAN model makes it suitable for healthcare applications requiring timely feedback, such as rehabilitation. 

The main contributions of this paper are summarized as follows: (1) A novel Multi-Residual Spatio Temporal Graph Network (MRSTGN) that incorporates both angular and positional 3D skeletons for patient action assessment; (2) A multi-residual architecture that captures the spatio-temporal dynamics of patient actions by integrating information from multiple residual layers, each of which extracts features at different levels of abstraction; (3) An attention fusion mechanism that selectively combines the information from the angular and positional skeletons, allowing the network to adaptively weight the importance of different features and providing a more comprehensive and accurate assessment of patient action; (4) Guidance: The system provides clear guidance on which body parts or movements to focus on and has shown high performance on physical rehabilitation benshmarks; (5) High performance: The efficiency of the proposed model is shown through extensive experimentation on  physical rehabilitation datasets UI-PRMD.

\section{Related Works}
\label{sec:related}
%(removed by Rim)Rehabilitation exercises play a crucial role in the recovery process for patients with various health conditions. 
%Evaluation of the quality and effectiveness of rehabilitation exercises is an essential task for healthcare professionals to ensure that patients receive optimal care.
3D skeleton-based approaches have emerged as a promising method for evaluating patient rehabilitation exercises due to their ability to capture the patient's posture and movement accurately. There are several approaches for evaluating patient rehabilitation exercises based on 3D skeletons. 

\subsection*{Discrete Movement Score (DMS) approaches}
These approaches involve quantifying specific movements performed by patients during rehabilitation exercises and assigning a score based on the quality of the movement. One example of a DMS approach is the Fugl-Meyer assessment \cite{gladstone2002fugl}, which is commonly used to evaluate the motor function of stroke patients. The Fugl-Meyer assessment consists of a series of tasks, each of which is scored based on the quality of the patient's movement. Other examples of DMS approaches include the Motor Assessment Scale, the Action Research Arm Test, and the Modified Ashworth Scale. DMS approaches have been shown to be reliable and valid for evaluating patient rehabilitation exercises. A study by Abreu \emph{et al.} ~\cite{abreu2017assessment} found that the Fugl-Meyer assessment had high inter-rater reliability and good validity when used to evaluate stroke patients. Another work by Van der \emph{et al.} ~\cite{van2001responsiveness} suggests that the Motor Assessment Scale was a reliable and valid measure of motor function in stroke patients. However, in general, DMS approaches have numerous limitations including subjectivity in scoring, leading to inter-rater variability; limited assessment scope, focusing on specific movements rather than overall functional abilities; time-consuming and resource-intensive process, requiring trained professionals; limited generalizability to populations beyond the specific target group.

\subsection*{Template-based approaches}

Template-based approaches are used to evaluate the performance of rehabilitation exercises performed by patients by comparing them to pre-defined templates or models of correct movements. 
One example is the GAITRite system \cite{bilney2003concurrent}, which is used to evaluate gait performance in patients.

Other examples of template-based approaches \cite{henderson2007virtual} include the Virtual Peg Insertion Test and the Virtual Reality Rehabilitation System. A study by Menz \emph{et al.} \cite{menz2004reliability} propose the GAITRite system to measure of gait performance in patients with Parkinson's disease. Another study by Subramanian \emph{et al.}. \cite{subramanian2013arm} found that the Virtual Reality Rehabilitation System was a valid and effective tool for upper limb rehabilitation in stroke patients.
These approaches are limited by the lack of individualization, limited variability, and inflexibility to adaptations. They may struggle to capture complex movements accurately and have a narrow scope of assessment. These limitations highlight the need for considering alternative assessment methods to provide a comprehensive evaluation of patient progress and functional abilities during rehabilitation exercise
\subsection*{Deep Learning based approaches}

Deep learning approaches have been increasingly used in recent years to evaluate patient rehabilitation exercises. These methods provide a powerful framework for analyzing complex, multi-modal data from different sources, such as motion sensors, video, and physiological signals, to provide accurate and reliable assessments of patient performance. One of the most popular deep learning models used for this task is the convolutional neural network (CNN). CNNs can learn discriminative features from raw sensor data and classify different types of exercises based on their patterns. For instance, researchers have used CNNs to classify different yoga poses using data from wearable motion sensors \cite{anand2022yoga,wu2019yoga}. Another popular approach is to use graph convolutional networks (GCNs) to capture the spatial relationships between different parts of the body during exercise movements. GCNs can model the human body as a graph, where the nodes represent the body parts and the edges represent the connections between them. Researchers have used GCNs to predict the muscle activity of different body parts during squats \cite{song2021secure} and to detect the correct execution of sit-to-stand exercises in patients with Parkinson's disease \cite{guo2020sparse}.
The investigation of graph convolutional network (GCN) based approaches for rehabilitation holds great promise in advancing the field providing the ability to capture spatial relationships between body parts during exercise movements. This article proposes a novel GCN-based approach for accurate and real-time evaluation of rehabilitation exercises.

\section{Proposed approach}
In this section, we introduce MR-STGAN, our proposed approach that incorporates a multi-residual spatio-temporal graph with attention fusion mechanisms.

\subsection{Problem Formulation}

\begin{figure*}[h!]
\centering
\includegraphics[width=14cm,height=8.4cm]{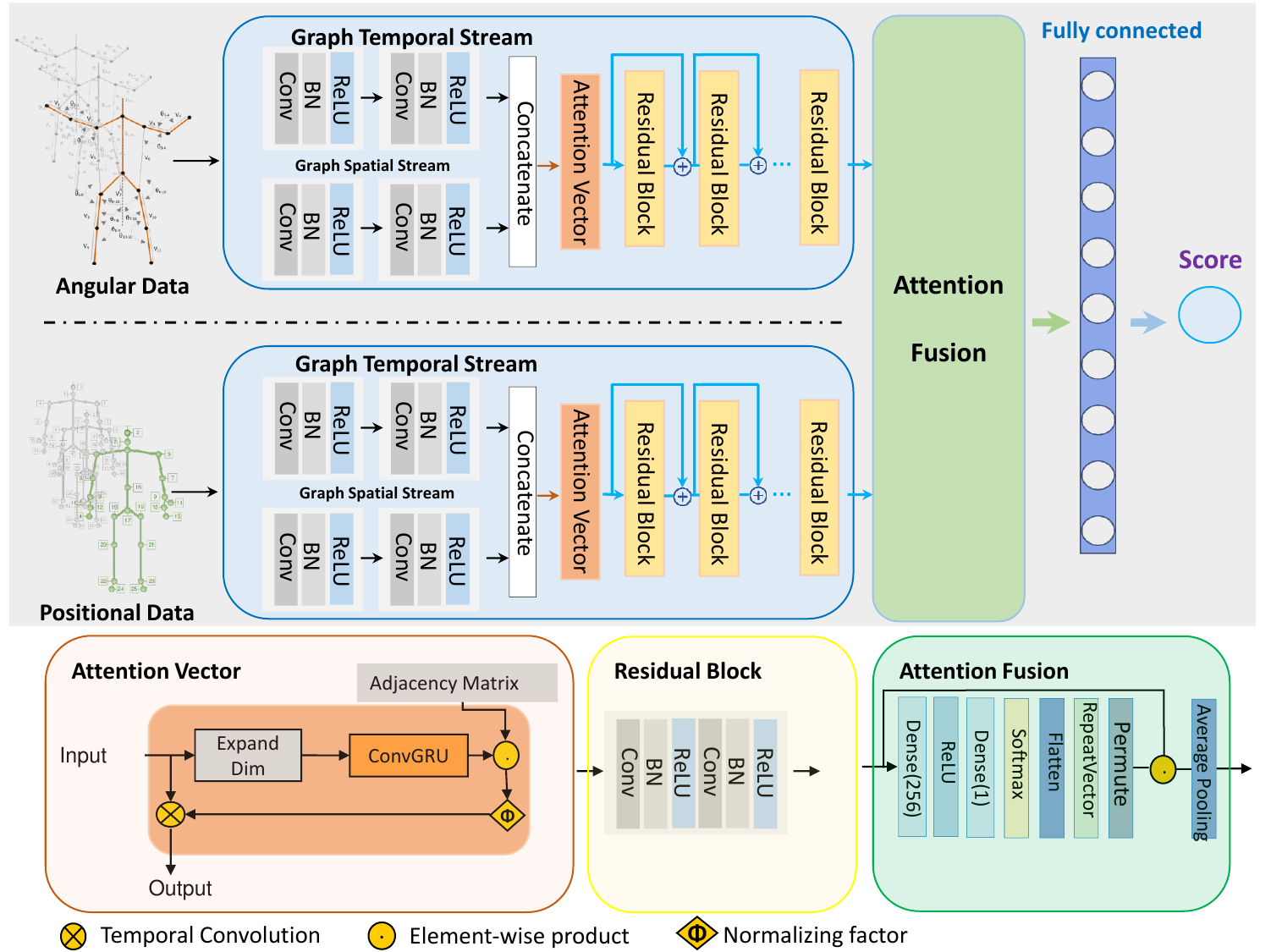}
\caption{Flowchart of the proposed approach.}
\label{fig:flowchart}
\end{figure*}

In this study, each exercise of rehabilitation is represented by an RGBD video denoted by $V_i = {X_{t=1...T}}$, where $X_t$ is the $t^{ith}$ frame, and a ground-truth performance score $y_i \in [min_{score},max_{score}]$. To ensure the robustness of the data to changes in body sizes, motion rates, camera perspectives, and interference backgrounds, skeleton-based data encoding is chosen, where each skeleton has $N$ joints with C-dimensional coordinates and angles. The goal is to predict the score $\hat{y}_j$ and provide feedback to the patient to improve their exercise fluency. The proposed MR-STGAN model predicts the ground-truth performance score and highlights articulations that require improvement to provide feedback to the patient. The flowchart of our proposed approach is depicted in Fig. \ref{fig:flowchart}. To ensure robustness to changes in body sizes, motion rates, camera perspectives, and interference backgrounds, we opted for skeleton-based data encoding over RGB image-based modality. Our encoding represents a sequence of skeletons, with each skeleton having $N$ joints, where each joint has C-dimensional coordinates. We encode these coordinates using a pose estimation approach or the sensor that captured the data. This encoding results in $V_i \in \mathbb{V}^{T \times N \times C}$ for each video and $X_t \in \mathbb{R}^{N \times C}$ for each frame. Our objective is to predict the score $\hat{y}_j$ for a given exercise, where each skeleton's motion plays a crucial role. We also aim to provide feedback to the patient to improve the fluency of their exercise. To achieve this, we use a self-attention vector $\chi^t \in \mathbb{R}^{T \times N \times 1}$, which identifies the joints that need improvement and highlights them for the patient.

\subsection{Multi-Residual Spatio Temporal Graph Network (MR-STG)}

The human skeleton can be represented as a directed acyclic graph, with joints as nodes and edges representing biomechanical dependencies. The graph structure representing each frame of the input skeleton sequence is denoted by $G = (N_G, E_G, A)$, where $N_G$ is the set of nodes and $E_G$ is the set of edges. The adjacency matrix of the graph is denoted by $A$. 

We preprocess both, positional and angular data by the MR-STG which consists of two streams: temporal and spatial. The temporal stream processes the input data $V$ as $T_s=K_a \otimes V$, where $\otimes$, and $K_a$ denote respectively the temporal stream convolution operations and the used kernel.

Similarly, the spatial stream processes the input data, but with different kernel sizes $S_s=K_a \otimes V$. The output of the two streams is concatenated $Z= T_s \OPLUS S_s$ where $\OPLUS$ denotes the concatenation operation. Next, we perform a graph convolution using the following equation: $G(Z) =(Z.A).W$. Where $W$ is a trainable weight matrix that is shared across all graphs to capture common features. For each kernel, $Z.A$ computes the weighted average of a node's features with those of its neighboring nodes, which is multiplied by $(Z.A).W$, a weight matrix. This operation helps to extract spatial features from the non-linear structure of the skeletal sequence.

Then, we use a Convolutional Gate Recurrent Unit (GRU) layer to compute a self-attention vector ($\chi^t \in \mathbb{R}^{T \times N \times 1}$) which makes the adjacency matrix dynamic and is recomputed each time through the added layer. Compared to RNN and LSTM, ConvGRU is low time-consuming due to its simpler structure with gates in one unit. The benefits of using ConvGRU instead of other layers in terms of computation time and accuracy on metrics were evaluated in the experimental results. ConvGRU combines CNN and GRU, maintaining the spatial structure of the input sequence and facilitating the extraction of spatial-temporal features.
For an action sequence ${x_{t=1...T}}$ with $T$ frames, forward propagation is performed as follows:

\begin{equation*}
z_t = \sigma(w_{zx}\otimes x_t + w_{zh}\otimes h_{t-1}+b_z)
\end{equation*}
\begin{equation*}
r_t = \sigma(w_{rx}\otimes x_t + w_{rh}\otimes h_{t-1}+b_r)
\end{equation*}
\begin{equation*}
o_t = tanh(w_{ox}\otimes x_t + w_{oh}\otimes (r_t \times f_{t-1}) +b_o)
\end{equation*}
\begin{equation*}
h_t = z_t\times x_t + (1-z_t)\times o_t
\end{equation*}

Here, $\otimes$ and $\times$ denote the convolution operation and Hadamard product, respectively. $tanh$ and $\sigma$ represent tangent and sigmoid functions. $w_x$, $w_h$, and $b$ are corresponding weights and biases. $h_t$ denotes the hidden state for each time index $t=1..T$ ($h_0$ is set to $0$) and is considered both the output and background information flowing into the network. The different gates in GRU are represented by $z_t$, $r_t$, and $o_t$. Finally, we inject an adjacency matrix $A$ by element-wise multiplication with the output of ConvGRU, and apply a normalization factor. The resulting output is then processed through a set of residual blocks, which consist of multiple residual layers. Finally, we average the output of the latter.

\subsection{Attention Fusion}
The use of Attention fusion between spatio-temporal angular and positional features ($F_{positional}$ and $F_{angular}$) is a significant contribution to the field of patient action assessment. By combining these two types of features, the proposed method is able to provide a more comprehensive understanding of the patient's movements during physical therapy exercises. To achieve this, the input tensor is split into two branches: one for the positional data and one for the angular data. 

 \begin{equation*}
  C = [F_{positional}, F_{angular}]   
 \end{equation*}

The concatenated tensor $C$ is then passed through a dense layer with units=256 and activation='relu', followed by another dense layer with units=1 and activation='softmax'. This results in a tensor of shape (batch, timesteps, 1) representing the attention weights for each frame.
 \begin{equation*}
Att = softmax(relu(C * W1 + b1) * W2 + b2)
 \end{equation*}

 where $W1$ and $W2$ are weight matrices, $b1$ and $b2$ are bias vectors, relu is the rectified linear unit activation function, * denotes element-wise multiplication and softmax function is used to ensure that the attention weights sum to one. The attention weights are then flattened using the Flatten layer $F$ and repeated $N_{features}$ times along the second axis using the RepeatVector layer. 

 \begin{equation*}
F = repeat(Att, N_{features})
 \end{equation*}
 
The tensor is then transposed using the Permute layer to have a shape of (batch, $N_{features}$, timesteps) to allow element-wise multiplication with the concatenated tensor: $D = C * F$.

Finally, the attended features $D$ are pooled using global average pooling with the GlobalAveragePooling1D layer to obtain a tensor of shape (batch, $N_{features}$), which represents the fused features for each example in the batch.
This method combines the advantages of both angular and positional features to better capture the nuances of human motion. The angular features provide information on the orientation of the body segments, while the positional features provide information on the spatial location of the body segments. 

By combining these two types of features, the model can better understand the relationship between the different body segments and their movements. The attention mechanism further enhances the model's ability to focus on the most informative parts of the input. The attention weights are learned during the training process and allow the model to selectively attend to certain parts of the input at different times. This means that the model can dynamically adjust its focus as the action progresses, which is particularly important for assessing complex actions. In addition, the use of attention fusion between the concatenated spatio-temporal angular and positional 3D skeletons features helps to address some of the limitations of previous methods. For example, previous methods \cite{deb2022graph} have relied solely on either angular or positional features, which may have limited their ability to capture all aspects of human motion. 

\section{Experiments}

This section evaluates the proposed MR-STGN model through comparative experiments using a rehabilitation exercise dataset and evaluation metrics. Implementation details, ablation studies, and quantitative comparisons with state-of-the-art approaches are also discussed.

\subsection{Datasets and Metrics}

The UI-PRMD dataset \cite{vakanski2018data} is publicly available and contains movements of common exercises performed by patients in physical rehab programs. Ten healthy individuals performed 10 repetitions of various physical therapy movements, captured using a Vicon optical tracker and a Microsoft Kinect sensor. The dataset includes full-body joint positions and angles, and its purpose is to serve as a foundation for mathematical modeling of therapy movements and establishing performance metrics to evaluate patients' consistency in executing rehabilitation exercises.

\subsection{Experimental Results}
The  MR-STGN model was implemented in Python 3.6 using the TensorFlow 2.x framework on a PC with an Intel Xeon Silver 4215R CPU, 32GB of RAM, and a GeForce GTX 3080 Ti 16GB RAM graphics card. The model was trained for 1500 epochs using the Adam optimizer and batch sizes of 4. The learning rate was set to $1e-4$. 
We utilized Huber loss in our study, benefiting from its robustness, balanced nature between MSE and MAE, and reduced sensitivity to outliers. The best model was selected based on its performance on the validation set and was evaluated on the test set. The used evaluation metric is the mean absolute error (MAE) metric.

We present the results obtained from various state-of-the-art approaches, such as the spatio-temporal neural network model, deep CNN model, and deep LSTM model as proposed in \cite{liao2020deep}. Additionally, we include the co-occurrence model \cite{li2018co}, PA-LSTM \cite{shahroudy2016ntu}, two-stream CNN \cite{simonyan2014two}, and hierarchical LSTM \cite{vakanski2018data}. Within our evaluation, we offer the findings derived from performing an ablation study on our model. This study involved using the same parameters with variations in data input, specifically focusing on positional data alone, angular data alone, and a combination of both.

The results in the table \ref{Tab:MetricsComparison} showcase the performance of various approaches, including our proposed method, across multiple exercises. Our approach, utilizing both positional and angular data, consistently demonstrates superior performance compared to other methods. When considering only positional or angular data, our approach outperforms other individual data-based approaches as well. Moreover, state-of-the-art approaches offer only limited performance for all exercise types presented in the dataset. It is worth mentioning that Exercise 10 (Standing shoulder scaption) is relatively easy to evaluate because it involves a single plane of motion and a few patient body joints. However, it is not perfectly assessed compared to our proposed approach, which provides a good assessment score for this exercise. The average error for our method is the lowest, indicating its effectiveness in accurately assessing rehabilitation exercises. 
The results from the time execution table \ref{Tab:time} highlight the speed of our model compared to Deb \emph{et al.} approach \cite{deb2022graph}. During the training phase, our model for the whole considered 134 videos taking only 29 minutes. Moreover, during the testing phase, our model demonstrated important speed, processing 34 exercise videos in just 3.68 seconds. These results underscore the advantage of our model in terms of both training and testing time, providing a more efficient and practical solution for evaluating rehabilitation exercises.

\begin{figure}[h!]
\centering
\includegraphics[width=3.8cm,height=4.5cm]{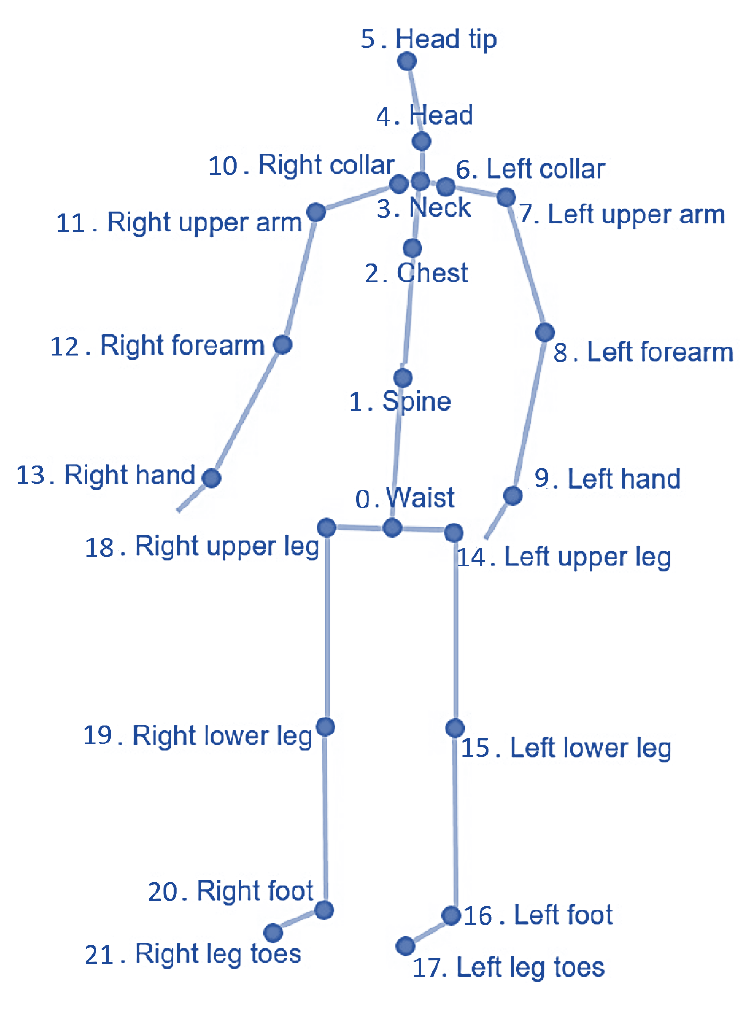}
\caption{Joints in the skeletal model from UI-PRMD dataset.}
\label{fig:skeletons}
\end{figure}

\begin{figure}[h!]
\centering
\includegraphics[width=8cm,height=6cm]{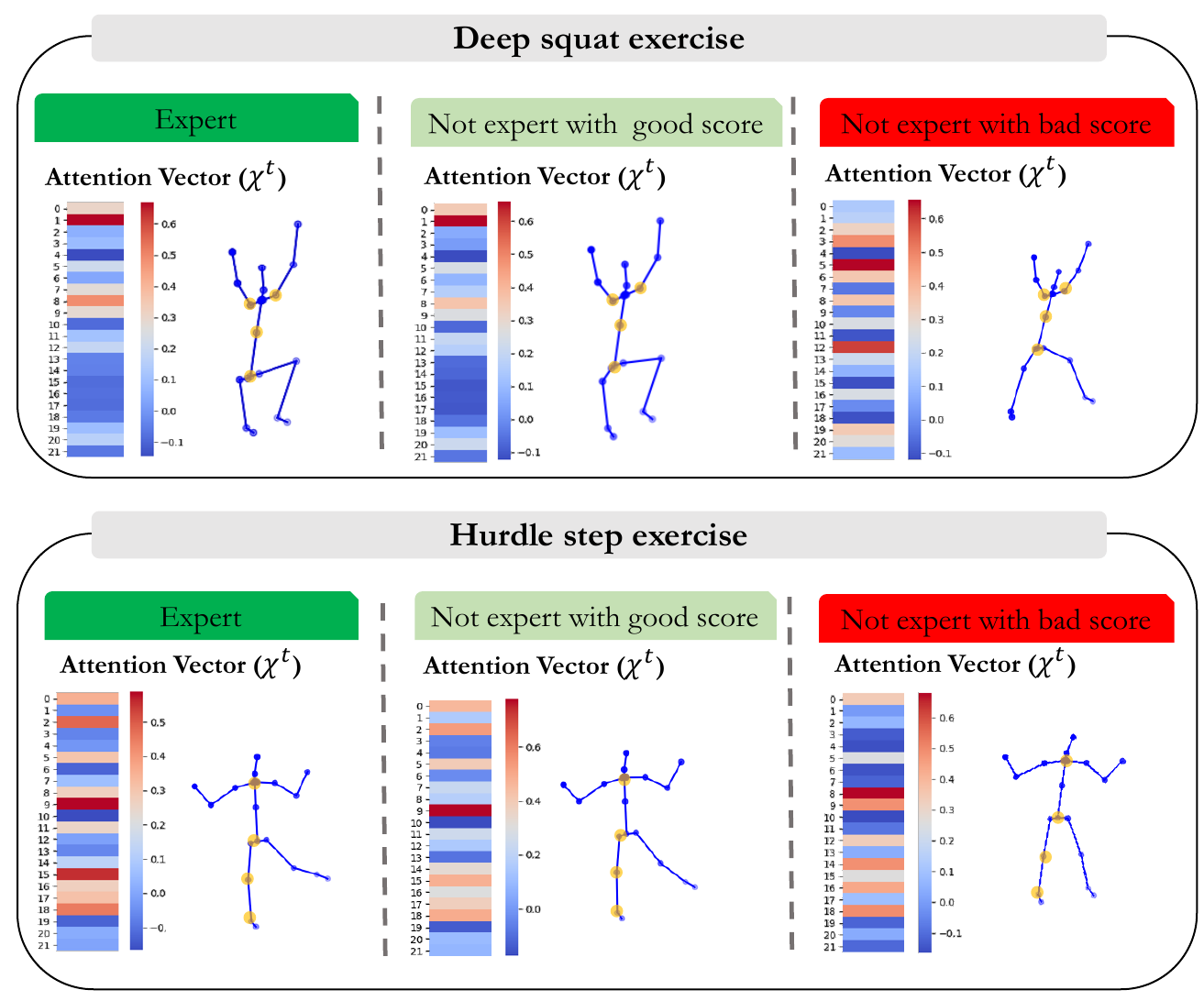}
\caption{Feedback visualization for different user profiles on UI-PRMD dataset: expert, not an expert with a good score, and not an expert with a low score.  The attention Vector denotes the joint role vector (hot colors represent high values).
Colored circles on the skeleton bodies allow the visualization of the attention vector and the role of body joints for different exercises.}
\label{fig:feedback}
\end{figure}

Fig. \ref{fig:feedback} demonstrates the computation of joint roles for non-expert users by the attention joint vectors $\chi^t$. It reveals that these roles differ from the expert's pattern when the patient receives a low assessment score. The visualization includes the Deep squat exercise (where the subject bends the knees to descend the body toward the floor with the heels on the floor, the knees aligned over the feet, the upper body remains aligned in the vertical plane) and the Hurdle step exercise (where the subject steps over the hurdle, while the hips, knees, and ankles of the standing leg remain vertical) along with their respective $\chi^t$. Refer to Fig. \ref{fig:skeletons} to identify the main joints considered for each exercise.

\begin{table*}[h!]
\centering
\begin{adjustbox}{max width=1\textwidth}
\begin{tabular}{p{1cm} p{1cm} p{1.2cm}p{1.2cm} p{1cm} p{1cm} p{1cm} p{1cm} p{1cm} p{1cm} p{1cm} p{1.5cm} p{1.35cm}} 
 \hline
Exercise  & Ours with both & Only positional data (ours)  & Only angular data (ours) &  Deb \emph{et al.} \cite{deb2022graph}  & Du \emph{et al.} \cite{du2021assessing} & Deep CNN \cite{liao2020deep} & Liao \emph{et al.} \cite{liao2020deep}& Deep LSTM \cite{liao2020deep}& Co-occurence \cite{li2018co}& PA-LSTM \cite{shahroudy2016ntu} & Hierarchical LSTM \cite{vakanski2018data}& Two-stream CNN \cite{simonyan2014two}\\
 \hline 
Ex1 & \textbf{0.008} & 0.012   &  0.011 & 0.012 & 0.009 & 0.013 & 0.010   &0.016  & 0.010 & 0.018 &0.030 &0.288 \\ 
Ex2 & \textbf{0.010} & 0.014  & 0.011   & 0.011 & 0.020 & 0.029 & 0.028  & 0.049 & 0.029  & 0.044 & 0.077 &0.223\\ 
Ex3 & \textbf{0.010}& 0.016  &  0.013 & 0.015 & 0.036 & 0.041  & 0.040  & 0.093 & 0.055  & 0.080 &0.137 &0.204\\ 
Ex4 & \textbf{0.008 }& 0.011  &  0.010 & 0.010 & 0.014 & 0.016 & 0.011  & 0.016 & 0.013  & 0.023 &0.035  &0.360\\ 
Ex5 & \textbf{0.007}& 0.013  & 0.014  & 0.010  & 0.014 & 0.013 & 0.018   &0.025  & 0.016  &0.031  &0.063  &0.123 \\ 

Ex6 & \textbf{0.010} & 0.015  & 0.018  & 0.017 & 0.020 & 0.023 & 0.017  &0.021  &  0.018 & 0.034 &0.046  &0.211  \\ 
Ex7 & \textbf{0.020}  &  0.021 & 0.025  & 0.023 &0.021 & 0.033 & 0.038 &  0.041  &0.027  & 0.049 & 0.192 & 0.050\\ 
Ex8 & \textbf{0.020} & 0.026  & 0.021  & 0.024 &0.022 & 0.029 & 0.023 &  0.046  &0.024  &0.050  &0.072  &0.043\\ 
Ex9 & \textbf{0.014} & 0.017  & 0.019  & 0.017 &0.025 & 0.024  & 0.022 &  0.044 &0.027  & 0.043 &0.065  &0.144\\ 
Ex10 & \textbf{0.015}  & 0.026  & 0.028  & 0.025 & 0.026 & 0.036 & 0.041  &0.052  &0.046  & 0.077  &0.160  &0.110 \\ 
\hline 
%\hline
Average & \textbf{0,012} & 0,017 &  0,017 & 0,016 & 0,021 & 0,026 & 0,025  & 0,040 &0,027   & 0,045 & 0,088&0.175 \\ 
\hline
\end{tabular}}
\end{adjustbox}
\caption{
Results of our method in comparison with state-of-the-art approaches on the UI-PRMD dataset.}
\label{Tab:MetricsComparison}
\end{table*}

\begin{table}[h!]
\centering
\begin{adjustbox}{max width=0.6\textwidth}
\begin{tabular}{c c c c c} 
 \hline
&Phase & Number of Videos  &  Execution Time\\
 \hline
Deb \emph{et al.} \cite{deb2022graph}&Train & 134 &  22 hours  \\ 
Our model & Train & 134 & \textbf{21 min}\\ 
 \hline
Deb \emph{et al.} \cite{deb2022graph}&Test & 34 & 13.87 seconds  \\ 
Our model & Test & 34  & \textbf{3.68 seconds}\\ 
 \hline
\end{tabular}
\end{adjustbox}
\caption{Computational time for Ex5 of UI-PRMD dataset for 1500 epochs.}
\label{Tab:time}
\end{table}


\section{Conclusion}
In summary, this paper introduces a novel approach called MR-STGAN, which utilizes 3D skeletons and graph convolutional networks for evaluating patient actions in physical rehabilitation exercises. The proposed model incorporates a multi-residual architecture to capture the spatio-temporal dynamics of patient actions, leveraging information from multiple residual layers. An attention fusion mechanism is employed to selectively combine angular and positional skeleton data, allowing for adaptive weighting of different features and personalized feedback to patients. The performance evaluation on the UI-PRMD dataset demonstrates that the proposed model surpasses existing state-of-the-art models in terms of quantitative measures. Furthermore, the model provides valuable guidance on specific body parts and movements. Its real-time performance makes it suitable for healthcare applications requiring timely feedback. Future work aims to explore continuous assessment possibilities and develop a graphical interface for visual feedback to patients.

\bibliographystyle{IEEEtran}
\bibliography{egbib.bib}

\end{document}